\documentclass[a4paper,12pt]{ieeeaccess}
\usepackage{mathptmx}
\usepackage{cite}
\usepackage{amsmath,amssymb,amsfonts}
\usepackage{algorithmic}
\usepackage{graphicx}
\usepackage{setspace}
\usepackage{textcomp}
\usepackage[margin=1in,includefoot,includehead]{geometry}
\def\BibTeX{{\rm B\kern-.05em{\sc i\kern-.025em b}\kern-.08em
    T\kern-.1667em\lower.7ex\hbox{E}\kern-.125emX}}
\begin{document}
\singlespacing
\history{Date of publication 2021/18/10, date of current version xxxx 00, 0000.}
\doi{10.1109/ACCESS.2021.DOI}

\title{Analysis of ECG to detect Afib in Healthgauge Watch}
\author{\uppercase{Arjun SK}, 
\uppercase{Sai Bhargav, and Rahul Guntha}}
\address{Dept of Computer Science, University of Alberta}

\markboth
{Author \headeretal: Preparation of Papers for IEEE TRANSACTIONS and JOURNALS}
{Author \headeretal: Preparation of Papers for IEEE TRANSACTIONS and JOURNALS}

\begin{abstract}
Atrial fibrillation(termed as AF/Afib henceforth) is a discrete and often rapid heart rhythm that can lead to clots near the heart. We can detect Afib by ECG signal by the absence of p and inconsistent intervals between R waves as shown in fig(1). Existing methods revolve around CNN that are used to detect afib but most of them work with 12 point lead ECG data where in our case the health gauge watch deals with single-point ECG data. Twelve-point lead ECG data is more accurate than a single point. Furthermore, the health gauge watch data is much noisier. Implementing a model to detect Afib for the watch is a test of how the CNN is changed/modified to work with real life data.
\end{abstract}


\titlepgskip=-15pt

\maketitle

\section{Introduction}
\label{sec:introduction}

Atrial Fibrillation is a condition of the heart that may go undetected for years unless a heart specialist checks the ECG report of the patient in question. AF affects about 1\% of the population \cite{zoni2014epidemiology}).
There are around 2.3 Million cases in the US. For early detection and decrease in cases the ECG signals can be analyzed using advanced AI techniques.  ECG signals are essentially time-series data that can be analyzed. We can detect AF by spotting the irregularities in the ECG signal. Figure 1 shows a normal ECG signal. The letters P, Q, R, S and T correspond to the different parts of the heart: atriums and ventricles through which the electric signals pass on.\\
The wave notion for AF is characterized by:
\\
1) Absence of p waves.
\\
2) The length between the R points are not same throughout the signal.\\
Both of these conditions can be seen in Figure 2 which depicts the ECG signal for AF.
The challenges in determining AF through ECG analysis are:\\
1) ECG signals include random, low-frequency, and susceptible, resulting in the diagnosis results being unstable. \cite{wu2020study}.\\
2) Manual analysis of the ECG signals are subjective and susceptible to inter observer vulnerabilities since non AF signals can also exhibit irregular RR intervals and absence of P waves \cite{nurmaini2020robust}.

% \PARstart{T}{his} document is a template for \LaTeX. If you are 
% reading a paper or PDF version of this document, please download the 
% electronic file, trans\_jour.tex, from the IEEE Web site at \underline
% {http://ieeeauthorcenter.ieee.org/create-your-ieee-article/}\break\underline{use-authoring-tools-and-ieee-article-templates/ieee-article-}\break\underline{templates/} so you can use it to prepare your manuscript. If 
% you would prefer to use LaTeX, download IEEE's LaTeX style and sample files 
% from the same Web page. You can also explore using the Overleaf editor at 
% \underline
% {https://www.overleaf.com/blog/278-how-to-use-overleaf-}\break\underline{with-ieee-collabratec-your-quick-guide-to-getting-started}\break\underline{\#.xsVp6tpPkrKM9}

% If your paper is intended for a conference, please contact your conference 
% editor concerning acceptable word processor formats for your particular 
% conference. 

% IEEE will do the final formatting of your paper. If your paper is intended 
% for a conference, please observe the conference page limits. 

\section{Literature Review}
Artificial intelligence has made a lot of progress over the last few years especially. A lot of diverse fields have adapted AI methods for their benefits. We are highlighting some of the achievements in AI in these diverse fields below. AI has repeatedly proved its use in aiding the process of learning.

 \cite{lopez2019improving} harvests the importance of sensitivity, sustainability, intelligence and social service to implement a technology platform to improve the attention span of school children in Mexico. The Soft banks NAO robot is used here.  This humanoid robot can sit, walk, drag among many of its capabilities including speech recognition. This robot implemented in a elementary school in Mexico was found to be effective. Another instance of AI in learning is by Corey et al \cite{heath2019using}, PRT (Pivotal response treatment) is an effective tool. To ensure an efficient training for the caregivers, a concise UI is needed.In this paper, clinicians evaluated a UI build prototype as an aspect of a particapatory design framework. Initially, the facilitator notices the child to identify an action in which the child is engaged. The facilitator must then incorporate oneself into the action. This allows the facilitator to pause the activity in attempt to gain the child's attention. After gaining the child's attention, the facilitator can offer a clear instruction at the child's vocal tier. After that, the child is expected to react. If the response is a reasonable attempt given the child's vocal ability, the interventionist allows the child to immediately resume the previous activity as a reward. Using this technique has been experimentally proven to help children with ASD to develop vocal communication skills. The data was used to train a classification model based on three class labels: attentive, inattentive, and shared attention. PyAudioAnalysis is used to process audio. For the classification tasks, two SVM models were used, one trained to separate speech from noise and the other to distinguish between child and adult vocalisations. This methodology resulted in an overall classification accuracy of 79 percent.\cite{licona2019towards} The purpose of this paper is to present initial work on developing an experimental end-user assessment for dual-user haptic systems for hands-on training. Such systems combine the benefits of haptic computer-based training systems with those of supervised training, in which a professional trainer vigorously assists in the process of learning. The goal of this project is to evaluate other architectures, in order to draw relative conclusions about the benefits and drawbacks of each.Dexterous manipulation is required in many hand motion professions. In medicine, we use Computer-Based Simulators (CBS) with 3D models to teach students about anesthesia and suturing. These systems include a haptic interface that serves as the master and a software architecture that connects the master to the slave. The purpose of this paper was to present preliminary efforts on building an experimental end-user analysis for dual-user haptic systems for hands-on training. It was put through its paces with the ESC architecture. The initial results were not convincing, but they did allow for the identification of several technical and organisational issues that must be addressed in the near future.\\
 \hspace{1cm} While it can seem that AI is only making strides in fields wherein there are less risks involved, a lot of research has gone into making AI more trustworthy. One of the high risk areas wherein AI is used is for Air traffic control. Meghna et al \cite{singh2005visual}. This paper proposed a novel method for recognizing hand gestures for guiding air crafts on the runway by air marshals. For generating maxima the Radon transform (RT) is used. The projections of the RT transform is computed on the angle axis. To derive static motions, K means is applied.
 \cite{panchanathan2019interdisciplinary} came up with a novel framework combining AI and IOT to form a solution based on smart cities. They provided two frameworks as of concept for implementing the framework in specific sectors. The first is an indoor localization and information discovery solution aimed at visually impaired residents, while the second is the installation of Virtual Reality (VR) sports stadium attendance in a smart stadium test environment. The goals of smart city development should apply directly toward the citizen. Person-Centered Multimedia Computing (PCMC) is a multimedia design paradigm that revolves around the individual "person" and uses adaptive design features to meet the needs of the general audience. This proposes that the most successful means of addressing the smart cities challenge is to take a multidisciplinary strategy that combines different disciplines.
\begin{itemize}
\item Technology, Society, and the Environment from a Multidisciplinary Perspective
\item Citizen-Centered Design, Smart City Infrastructure and Dynamics, and Socio-Environmental Practices and Policies are the three research thrusts.
\end{itemize}
Interdisciplinary Perspectives:
\begin{itemize}
\item Technology: IOT and IOC
\item Society: We should make sure that smart city study complies with societal ethical, safety, and cultural standards.
\item The Environment: Smart City solutions can also be assessed for their long-term viability, environmental effect, scalability, and usability.
\end{itemize}
An interesting use case of AI in gaming is present in \cite{fakhri2019foveated}.
This is an approach for depicting visual information through haptics for 3d environments. This approach is tested using the game Doom which is a first person shooter game. The study presented an encourage result for blind as well as sighted people.

The most notable contributions that have benefited the healthcare industry are  
Rossol et al  \cite{rossol2011framework} demonstrated the ability to develop training modules for the severely disabled to use wheelchairs through VR. They used Bayesian Networks for self adjusting adaptive training. The major issues they addressed using their solution were the power and cost of the wheelchair which got reduced to a bare minimum. The framework ensured addressing the endurance of a range of patient skill's so that the training is effective in nature.

\cite{ponce2019implementing}
is another notable paper wherein robotics and AI are combined to solve problems in medicine. In this paper, the authors implement a robotic platform (RP) for therapies in Mexico.  Considering Mexico is an developing country the authors considered factors along technology, culture and economy. The technological factors they considered deals with the pressure sensor for interacting with children, image processing for detecting distance and facial expression of children, robots facial movements and robots body movements. Meanwhile the cultural factors involved relate to parents not being familiar with robots. therapists not accepting the robot because they prefer conventional tools, hospitals not familiar with technology and community not familiar with robots. The economic factors correspond to parents, social programs and therapists requiring a low price for the rebound design of robot should be cheap. With all these in consideration the novelty of the RP comes in picture.

\cite{mourey2019human} is another paper that focuses on the human body fall detection using various stages of detection techniques which are video analysis, Body Recognition, and Trigger alert. OpenCV is used to implement the algorithm. Testing of the algorithm has been done on various lightning and environmental conditions.
As fall detection might cause serious injuries to the human body, a technique has to be developed to detect the fall either by analyzing the video in each frame from the background. Though this can be developed by hardware that involves pressure acceleration and sound sensors to detect the fall a person needs to always wear a hardware device. But using video-based system procedures which involve video Acquisition, Analysis, and Notifying the user about the fall detection. With this algorithm, we can also analyze the severity of the fall by checking if the subject is under stable condition or not.
\begin{enumerate}
\item Sensor-based method: Detection is inferred by the embedded sensors that sense the movement and position of the object and also use external sensors including pressure and acoustic sensors.
\item Vision-based Technique: Shape analysis, shape deformation, and posture estimation can be used to describe the human shape which is approximated by the ellipse. Bounding box and ellipse approximation can be both used for best results. Regarding classification of body posture, Support vector machine or Multi-class SVM classification algorithm can be used.
\item Machine learning method: Fall can be classified based on various other activities such as bending, stripping, and tripping. We can use Support Vector Machines, Decision tree classifications for pattern classification. Bayesian belief networks and ANN can also be used to classify falls from daily activities.
\end{enumerate}
Le2i data set videos are used to research and predict a fall based on the human body's response to a sudden environment. This data set is not only limited to falling but also has images of sitting, resting, and bending.
The following stages are used to classify and detect the subject fall:
\begin{enumerate}
\item	Video Analysis: Convert the videos into frames. GMG and MOG2 methods are used for background subtraction and are triggered based on the motion estimation and illumination threshold. At the end of this step, the algorithm creates a binary mask frame of the moving object in the video.
\item	Human Body Approximation: Select the region of interest. The shape measurements are used to analyze and compare the defined thresholds to differentiate between humans and other objects.
\item	Fall Definition: Based on the major axis and floor and the size of the ellipse or the ratio of the width to the height of the bounding box exceeds the threshold it is considered as a fall.
\item	Fall can be detected in 3 scenarios:
\begin{itemize}
\item	Change of w/h
\item	Change of theta and ‘F’.
\end{itemize}

In some cases, both methods can be used to define the fall.

\item	Fall Detection: The algorithm will record the frame details and the absolute parameter values used to detect the fall.
\item	Detection, and Notification: After the fall has been detected we register the fall period and a notification is sent to the subject if the fall period is more than 5 frames.
\end{enumerate}
The Results are as foll:\\
	Method 1: The precision of the method is measured using (MAP)\\
	Method 2: We calculate the accuracy of the approach in comparison with the annotation file.\\
By following Method 1 it gave an accuracy of 85% whereas method 2 is 86.6%.
Challenges and Limitations :
\begin{enumerate}
\item	True detection of the subject (human)
\item	Differentiating between human activities to differentiate a fall.
\item	Creating an up-to-date notification system.
\end{enumerate}
\cite{ke2019race} differentiates races such as human and non-human eyes. The iris is a huge section of the human eye's ring between the pupil and the scalar with a lot of different textures. Because of the stability and specificity of its textural features, iris segmentation can help with problems like racial classification. However, The current method of racial categorization by iris image primarily relies on manual feature extraction and classification study, which has some drawbacks such as segment heterogeneous iris images with active contour model and prior noise characteristics. The model that has been finalised to perform the race classification is LGBP-SVM (local Gabor binary pattern (LGBP) and support vector machine (SVM). It begins by 
\Figure[t!](topskip=0pt, botskip=0pt, midskip=0pt){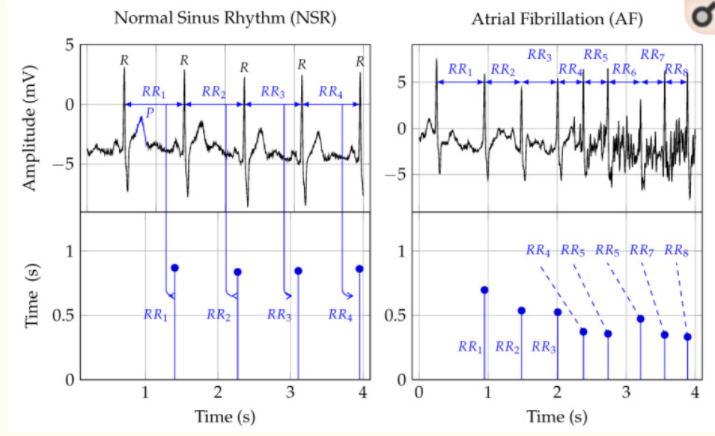}
{Absence of p waves and inconsistent intervals between R waves.\cite{faust2020review}\label{fig1}}
\Figure[t!](topskip=0pt, botskip=0pt, midskip=0pt){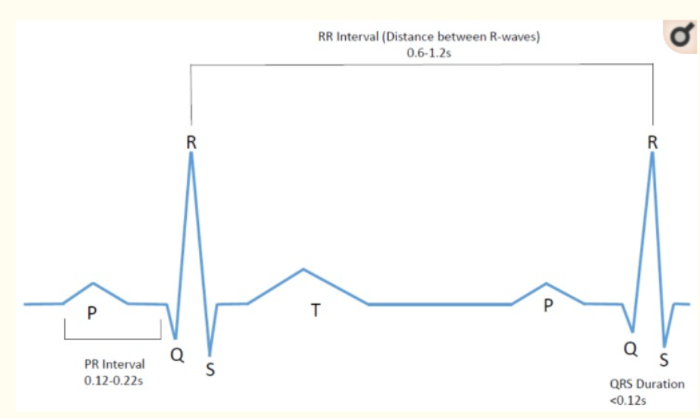}
{Absence of p waves and inconsistent intervals between R waves.\cite{ahmed2020early}\label{fig2}}
extracting textural information from iris images and categorising them into two groups. Furthermore, the active contour model based on the localising region and the circular Hough transform are used individually to localise iris regions using distinct methodologies. A variety of experiments have been conducted, including human vs. lions and Asians vs. whites, and they have confirmed that it enhances segmentation accuracy and proper segmentation rate. A variety of tests have been conducted, including human vs. lions and Asians vs. whites, and they have confirmed that it enhances segmentation accuracy and proper segmentation rate. In future, Researchers will also focus on multi-race iris categorization and the building of an effective and efficient iris recognition system in the future.\\
\cite{graham2019accurate}
The  KiTS19 data set is used for kidney segmentation using deep learning. A Sørensen–Dice score of 0.962 is achieved for this model. U net is used to section biomedical pictures of different modalities, including X-ray and CT checks.\cite{mcdaniel2020smart} focuses on the evaluation of roughness perception and fingertip vibration of a subject by performing a regression analysis. A tactile Mel filter bank has been created using the Pacinian corpuscle sensitivity model and tactile Mel frequency spectrum coefficient(TMFCC), which is an evaluation method for Mel frequency spectrum coefficients.
To optimize the Pacinian corpuscle sensitivity model and the tactile filter bank a sequential Model global optimization is performed. Using a fast Fourier transform on each part of the data a tactile filter bank is created which is superimposed on the calculated frequency spectrum and converted to a powered spectrum which is a time signal. On the time signal, a discrete cosine transformation is performed to obtain TMFCC features and the average obtained over each section is used as TMFCC feature value.
\cite{west2019assessing}

The results observed were:
\begin{itemize}
\item The newly obtained Pacinian characteristics have a similar tendency to the previously reported characteristics.
\item Based on this model, the created Mel filter bank the sensitivity of vibration is in the range of 200 Hz – 400Hz due to the Pacinian corpuscle’s characteristics.
\end{itemize}

\cite{west2019assessing}
The paper focuses on the capability of neural networks using deep learning models for assessing  Parkinson’s which is a progressive age-related neurogenerative disease that affects the motor movements of the human body. Whenever a patient is diagnosed with PD they lose 50–70\% of their dopaminergic neurons. As PD is ostensibly incurable, few treatments such as Exenatide may be effective during the early stage of the detection. Earlier MRIs are used to diagnose PD. But due to the traditional poor ability of the sMRI’s to detect subtle physiological changes in the areas that are associated with PD. In this paper, a total of four deep-learning-based models are proposed and multiple models are constructed to classify the patients using the sMRI’s data. We classify patients based on biomarkers found in the images of structural magnetic resonance and demonstrate high efficacy in the task of diagnosing Parkinson’s disease using the Convolution-Neural-Network model based on two approaches with 3D and 2D neural networks. 
Parkinson’s Progression Markers Initiative (PPMI) public dataset is used for data and preprocessing. The data needs to be standardized due to the morphological and dimensional differences between the scans. With the RAW data after the pre-processing images of the processing brain are obtained which are useful for the PD diagnosis. The models built gave an accuracy of 75\% and 76\% sensitivity. Of all the 4 models implemented the best performing 3D model was the 3D-CNN, with 75\% accuracy and balanced metrics in sensitivity, specificity, and precision. This model is promising as it suggests learning meaningful features rather than guessing a random class.\\
\cite{han2019level}
Image segmentation is one of the emerging areas of computer vision especially in the area of medical diagnosis. The article proposes a new level set method (LSM) with an adaptive hybrid region-based energy to achieve the highest accuracy during the image segmentation. 
In the first stage median region intensity are computed using the filter on an input image with traditional mean region intensity descriptions to design a novel global region-based signed pressure force (GRSPF) then global region-based energy Is defined using it is similar to that a new local region-based SPF (LRSPF) is also designed and the local region-based energy is defined using this LRSPF, to enhance the model’s versatility.  An adaptive weight to control the roles of the global and local region-based energies is introduced to construct the hybrid region-based energy, which drives the level set more appropriately. Using this technique gives more accurate results than existing level set methods. To validate the accuracy and effectiveness of the proposed LSM method outperforms on different sets of images used to which results are compared with GRSPF and LRSPF based LSM’s.\\
\cite{ma2019remote} proposes a solution to make the PPG sensor remote using a camera , advanced computer vision techniques and signal processing techniques. The authors also went a step
further to analyse the differences between industry grade camera and grayscale camera.  The facial image is converted to grayscale iomage for further signal extraction and heart rate detection
For the grond truth analysis, a BTChoicTM blood oxygen and dynamic heart rate bracelet was used.
They observed that make up, sunscreen, large amounts of facial hair and head movements lead to fluctations and inconsistencies in results. The future work here lies in improving computer vision techniques over these limitations

A major problem in medicine is of detecting Afib from ECG waves.
This is a problem at the intersection of biology and statistics due to which a lot of methods have already been developed. We have found the following types of methods:

\begin{enumerate}
\item Simple statistics based methods:
\cite{ahmed2020early}
In this paper Afib is detected by testing if the RR interval lengths change or not. To implement this a consecutive difference between time values of R peaks are calculated. 

Further the max and min of these differences are checked to see if the value is between 6 - 12s (normal heart)

\item Neural network based methods
The below papers are the popular neural networks based methods for detecting Afib. The idea is similar across all of them. The CNN is used for feature extraction and learns the threshold of values that is ideally seen in Afib.

We have highlighted the methods, dataset and the accuracy below for each of these papers

\begin{enumerate}
\item
\cite{hsieh2020detection}
Dataset: PhysioNet Challenge 2017
Algorithms: 1D CNN and length normalization algorithm for the dataset
Accuracy: f1 score - 78% 
\item
\cite{ping2020automatic}
Dataset:  Computing in Cardiology Challenge 2017
Algorithms: 8 layer CNN (with shortcut connection)  + 1 layer LSTM-> 8CSL 
Accuracy: Average 85%

\item
\cite{ma2020automatic}
Dataset:  MIT-BIH Atrial Fibrillation Database
Algorithms: CNN LSTM
Accuracy: 97.21%

\item
\cite{ullah2021hybrid}
Dataset: MIT-BIH
Algorithms: 1D and 2D model
Accuracy: 97.38%
\end{enumerate}

\end{enumerate}

\section{Project plan}

\subsection{Novelty}
The problem with all these neural network based papers are:

\begin{enumerate}
    \item They usually work on datasets that have been filtered and cleaned to an extent that the data does not resemble a real time ECG. Since a real time ECG developed from the Healthgauge watch is very noisy. This watch detects ECG only when a button is pressed which initiates contact of the watch with the skin. This contact is not perfect and thus the ECG waves that the watch generates are weak and noisy in nature. 
    \item The academic datasets used commonly are generated using a 12 point lead mechanism. A 12 point lead refers to 12 points of contact of the electrodes on the chest. Hence this data is very accurate. The data in the Healthgauge watch is generated using a single point of contact, hence a single lead. Further, the watch uses a PPG (Photoplethysmography) which is an optical measurement device. The PPG signals are a reflection of blood circulations and thereby ECG. ECG is computed from the PPG. Since the ECG is not the direct measurement the academic ECG 12 lead datasets can not be directly translated to our use case.

\end{enumerate}
Our solution revolves around converting the accurate dataset to a less frequency one and then training the model to get a better approximation. Another way is to learn using the high frequency data and then convert the frequency of the watch data so that both are on the same scale.

\subsection{Milestones and Breakdown}
\begin{enumerate}
    \item Literature review
    \begin{enumerate}
       \item Multimedia based methods: Arjun, Sai and Rahul: Oct 25
        \item CNN based methods: Arjun, Sai and Rahul: Oct 25
        \item Simple statistics methods: Arjun: Oct 25
    \end{enumerate}
    
      \item Deciding model parameters: Arjun, Sai and Rahul: Nov 5.
      \item Writing initial code to test on some data: Sai and Rahul: Nov 15
      \item Writing full code for the model: Arjun and Sai: Nov 30
      \item Documentation: Arjun and Rahul: Dec 5
\end{enumerate}

\section{Methods}
Regarding methods we implemented 2 different models 
\begin{enumerate}
    \item 1D CNN - Coming to 1D-CNN we developed a model containing 10 convolutional layers and each layer has an pooling layer. The architecture is given below Figure 3. Initially we shuffle the data set(MIT-BIH) to avoid bias while training and split the data set into 90\% and 10\% for train test split. Then before feeding the data into neural network we expand the dimension for 1D to 2D since CNN works on 2D. We are implementing 1D CNN since it is time series data. Coming to pooling layers we are taking Maxpooling layer since we need prominent features from graph.  We have used dropout layers to reduce over-fitting. Batch normalization is used to standardize the inputs and accelerate training. For backpropogation we are using Adam optimizer and binary cross-entropy loss function. The model is trained over 100 epochs and batch size of 128.
    \item 2D CNN - The important part for this model is to create a spectogram image data using time series data. Electrocardiograms are representations of the periodic cardiac cycle. They contain useful information both in the time and the frequency domain. We can extract frequency information by applying a Fourier transform and normalizing the frequency contributions to obtain the power spectrum of the ECG. we have implemented a 24 layer convolutional neural network to train our dataset over 50 epochs of batch size 50 and temporal aggregation is achieved using Lambda layer. The model achieved accuracy of 74\%.
\end{enumerate}
\Figure[t!](topskip=0pt, botskip=0pt, midskip=0pt){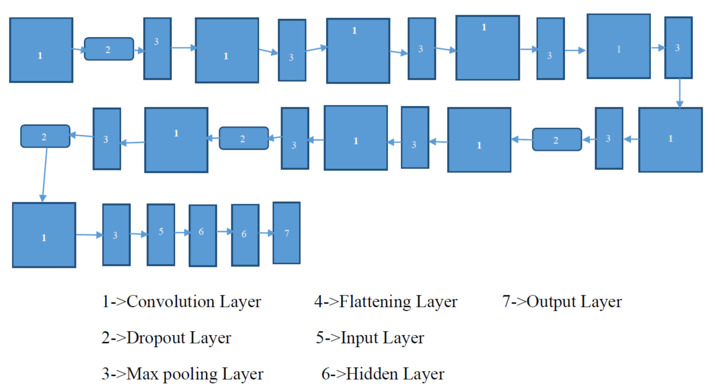}
{Architecture of the 1D CNN}
%{arc}
\section{Results}
Our results for both 1D CNN and 2D CNN have been surprising. While we expected the 2D CNN to work better than the 1D CNN, the results were opposite of that. For 1D CNN we got an accuracy of 82 \%. While the 2D CNN got an accuracy of 74 \%. As you can see in Figure 5, the accuracy of the 1D CNN is shown. Figure 6 shows how the loss changes over epochs. Training loss decreases over epochs while validation loss stagnates. For the 2D CNN training accuracy increases over epochs while the validation accuracy stagnates after sometime. The 2D CNN model detects outliers much better than the 1D model. But the accuracy is less which defeats the purpose.

\Figure[t!](topskip=0pt, botskip=0pt, midskip=0pt){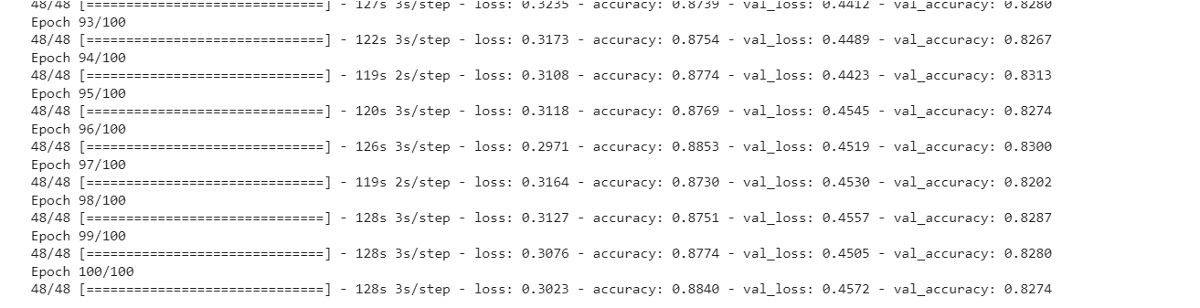}
{Accuracy of the 1D CNN}

\Figure[t!](topskip=0pt, botskip=0pt, midskip=0pt){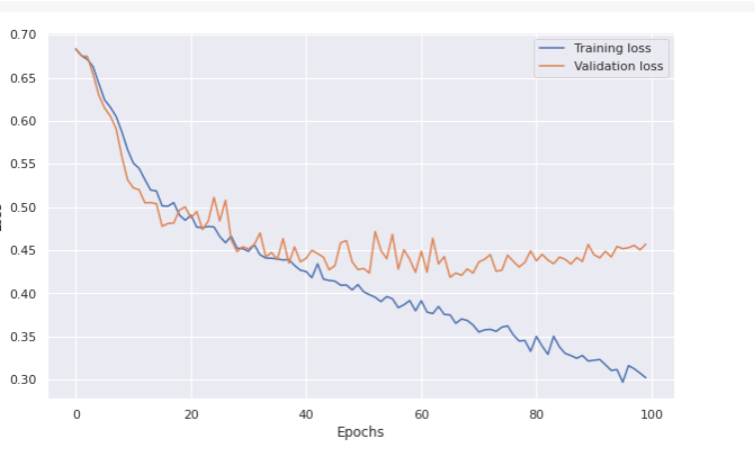}
{Loss VS Epochs for the 1D CNN}

\Figure[t!](topskip=0pt, botskip=0pt, midskip=0pt){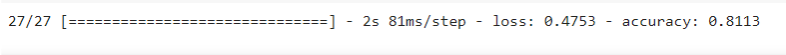}
{Accuracy of the 2D CNN}

\Figure[t!](topskip=0pt, botskip=0pt, midskip=0pt){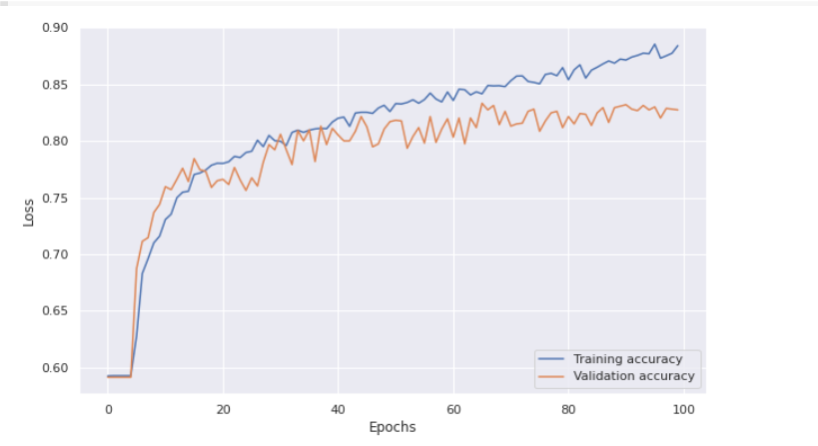}
{Loss vs Epochs for 2D CNN}

\section{Conclusion and Future work}
There are a lot of advancements in multimedia and especially research related to healthcare. Afib detection is one of them wherein a lot of strides have taken place over the last few years. Our methods may not have had the highest accuracy but they are novel and present future opportunities. They also present a valid comparison between 1D CNN and 2D CNN methods. Our observation is that 1D CNN methods perform better over 2D CNN methods. A reason for this also could be the 1D CNN works with original format of the data: a waveform time-series while the 2D CNN works with the image format essentially.

The main future work here is addition of layers and increasing epochs to increase the performance. Another change could be in the activation functions too. The most important future work we see is the conversion of the ECG wave into a single lead ECG wave. This is the most important improvement as doing this will result into a more relatable experience and can be used by all wearable devices. The 2D CNN's accuracy could be improved as this model deals with outliers very well. 

% Figure 1: \includegraphics{fig1.jpg}
% \cite{b9}

% Figure 2: \includegraphics{fig2.jpg}
% \cite{b4}

\bibliographystyle{plain}
\bibliography{bibliography.bib}

\begin{thebibliography}{10}

\bibitem{ahmed2020early}
Nuzhat Ahmed and Yong Zhu.
\newblock Early detection of atrial fibrillation based on ecg signals.
\newblock {\em Bioengineering}, 7(1):16, 2020.

\bibitem{fakhri2019foveated}
Bijan Fakhri, Troy McDaniel, Heni~Ben Amor, Hemanth Venkateswara, Abhik
  Chowdhury, and Sethuraman Panchanathan.
\newblock Foveated haptic gaze.
\newblock In {\em International Conference on Smart Multimedia}, pages
  132--144. Springer, 2019.

\bibitem{faust2020review}
Oliver Faust, Edward~J Ciaccio, and U~Rajendra Acharya.
\newblock A review of atrial fibrillation detection methods as a service.
\newblock {\em International journal of environmental research and public
  health}, 17(9):3093, 2020.

\bibitem{graham2019accurate}
John~Brandon Graham-Knight, Kymora Scotland, Victor~KF Wong, Abtin Djavadifar,
  Dirk Lange, Ben Chew, Patricia Lasserre, and Homayoun Najjaran.
\newblock Accurate kidney segmentation in ct scans using deep transfer
  learning.
\newblock In {\em International Conference on Smart Multimedia}, pages
  147--157. Springer, 2019.

\bibitem{han2019level}
Bin Han.
\newblock Level sets driven by adaptive hybrid region-based energy for medical
  image segmentation.
\newblock In {\em International Conference on Smart Multimedia}, pages
  394--402. Springer, 2019.

\bibitem{heath2019using}
Corey~DC Heath, Tracey Heath, Troy McDaniel, Hemanth Venkateswara, and
  Sethuraman Panchanathan.
\newblock Using participatory design to create a user interface for analyzing
  pivotal response treatment video probes.
\newblock In {\em International Conference on Smart Multimedia}, pages
  183--198. Springer, 2019.

\bibitem{hsieh2020detection}
Chaur-Heh Hsieh, Yan-Shuo Li, Bor-Jiunn Hwang, and Ching-Hua Hsiao.
\newblock Detection of atrial fibrillation using 1d convolutional neural
  network.
\newblock {\em Sensors}, 20(7):2136, 2020.

\bibitem{ke2019race}
Xianting Ke, Lingling An, Qingqi Pei, and Xuyu Wang.
\newblock Race classification based iris image segmentation.
\newblock In {\em International Conference on Smart Multimedia}, pages
  383--393. Springer, 2019.

\bibitem{licona2019towards}
Angel~Ricardo Licona, Guillermo~Zamora de~la Pena, Oscar~Diaz Cruz, Arnaud
  Lelev{\'e}, Damien Eb{\'e}rard, and Minh~Tu Pham.
\newblock Towards a dual-user haptic training system user feedback setup.
\newblock In {\em International Conference on Smart Multimedia}, pages
  286--297. Springer, 2019.

\bibitem{lopez2019improving}
Edgar Lopez-Caudana, Pedro Ponce, Nancy Mazon, Luis Marquez, Ivan Mejia, and
  Germ{\'a}n Baltazar.
\newblock Improving the attention span of elementary school children in mexico
  through a s4 technology platform.
\newblock In {\em International Conference on Smart Multimedia}, pages
  525--532. Springer, 2019.

\bibitem{ma2020automatic}
Fengying Ma, Jingyao Zhang, Wei Chen, Wei Liang, and Wenjia Yang.
\newblock An automatic system for atrial fibrillation by using a cnn-lstm
  model.
\newblock {\em Discrete Dynamics in Nature and Society}, 2020, 2020.

\bibitem{ma2019remote}
Xiaocong Ma, Diana~P Tob{\'o}n, and Abdulmotaleb El~Saddik.
\newblock Remote photoplethysmography (rppg) for contactless heart rate
  monitoring using a single monochrome and color camera.
\newblock In {\em International Conference on Smart Multimedia}, pages
  248--262. Springer, 2019.

\bibitem{mcdaniel2020smart}
Troy McDaniel, Stefano Berretti, Igor~DD Curcio, and Anup Basu.
\newblock {\em Smart Multimedia: Second International Conference, ICSM 2019,
  San Diego, CA, USA, December 16-18, 2019, Revised Selected Papers}, volume
  12015.
\newblock Springer Nature, 2020.

\bibitem{mourey2019human}
Jannatul Mourey, Ava~Sehat Niaki, Priyanka Kaplish, and Rupali Gupta.
\newblock Human body fall recognition system.
\newblock In {\em International Conference on Smart Multimedia}, pages
  372--380. Springer, 2019.

\bibitem{nurmaini2020robust}
Siti Nurmaini, Alexander~Edo Tondas, Annisa Darmawahyuni, Muhammad~Naufal
  Rachmatullah, Radiyati~Umi Partan, Firdaus Firdaus, Bambang Tutuko, Ferlita
  Pratiwi, Andre~Herviant Juliano, and Rahmi Khoirani.
\newblock Robust detection of atrial fibrillation from short-term
  electrocardiogram using convolutional neural networks.
\newblock {\em Future Generation Computer Systems}, 113:304--317, 2020.

\bibitem{panchanathan2019interdisciplinary}
Sethuraman Panchanathan, Ramin Tadayon, Troy McDaniel, and Vipanchi Chacham.
\newblock An interdisciplinary framework for citizen-centered smart cities and
  smart living.
\newblock In {\em International Conference on Smart Multimedia}, pages
  107--122. Springer, 2019.

\bibitem{ping2020automatic}
Yongjie Ping, Chao Chen, Lu~Wu, Yinglong Wang, and Minglei Shu.
\newblock Automatic detection of atrial fibrillation based on cnn-lstm and
  shortcut connection.
\newblock In {\em Healthcare}, volume~8, page 139. Multidisciplinary Digital
  Publishing Institute, 2020.

\bibitem{ponce2019implementing}
Pedro Ponce, Edgar~Omar Lopez, and Arturo Molina.
\newblock Implementing robotic platforms for therapies using qualitative
  factors in mexico.
\newblock In {\em International Conference on Smart Multimedia}, pages
  123--131. Springer, 2019.

\bibitem{rossol2011framework}
Nathaniel Rossol, Irene Cheng, Walter~F Bischof, and Anup Basu.
\newblock A framework for adaptive training and games in virtual reality
  rehabilitation environments.
\newblock In {\em proceedings of the 10th international conference on virtual
  reality continuum and its applications in industry}, pages 343--346, 2011.

\bibitem{singh2005visual}
Meghna Singh, Mrinal Mandal, and Anup Basu.
\newblock Visual gesture recognition for ground air traffic control using the
  radon transform.
\newblock In {\em 2005 IEEE/RSJ International Conference on Intelligent Robots
  and Systems}, pages 2586--2591. IEEE, 2005.

\bibitem{ullah2021hybrid}
Amin Ullah, Shanshan Tu, Raja~Majid Mehmood, Muhammad Ehatisham-ul haq, et~al.
\newblock A hybrid deep cnn model for abnormal arrhythmia detection based on
  cardiac ecg signal.
\newblock {\em Sensors}, 21(3):951, 2021.

\bibitem{west2019assessing}
Christopher West, Sara Soltaninejad, and Irene Cheng.
\newblock Assessing the capability of deep-learning models in parkinson’s
  disease diagnosis.
\newblock In {\em International Conference on Smart Multimedia}, pages
  237--247. Springer, 2019.

\bibitem{wu2020study}
Mengze Wu, Yongdi Lu, Wenli Yang, and Shen~Yuong Wong.
\newblock A study on arrhythmia via ecg signal classification using the
  convolutional neural network.
\newblock {\em Frontiers in computational neuroscience}, 14:106, 2020.

\bibitem{zoni2014epidemiology}
Massimo Zoni-Berisso, Fabrizio Lercari, Tiziana Carazza, and Stefano
  Domenicucci.
\newblock Epidemiology of atrial fibrillation: European perspective.
\newblock {\em Clinical epidemiology}, 6:213, 2014.

\end{thebibliography}

\EOD

\end{document}